%% file: main.tex
\documentclass[10pt,twocolumn,letterpaper]{article}

\usepackage{authblk} 
\author[1]{Jae Joong Lee}
\author[1]{Bedrich Benes}

\affil[1]{Department of Computer Science, Purdue University, West Lafayette, USA}
\affil[ ]{\texttt{\{lee2161, bbenes\}@purdue.edu}}

\usepackage{wacv}              

\usepackage{graphicx}
\usepackage{amsmath}
\usepackage{amssymb}
\usepackage{booktabs}
\usepackage{nicematrix}
\usepackage{soul}
\usepackage[accsupp]{axessibility} 
\usepackage[pagebackref,breaklinks,colorlinks]{hyperref}
\affil[ ]{\texttt{\url{https://www.jaejoonglee.com/wacv25_rgb2point}}}

\usepackage[capitalize]{cleveref}

\crefname{section}{Sec.}{Secs.}
\Crefname{section}{Section}{Sections}
\Crefname{table}{Table}{Tables}
\crefname{table}{Tab.}{Tabs.}

\def\wacvPaperID{01437} 
\def\confName{WACV}
\def\confYear{2025}

\begin{document}
\input{macros.tex}
\title{ 
\name: 3D Point Cloud Generation from Single RGB Images
}


\maketitle

\input{src/00-abs}
\input{src/01-intro}
\input{src/02-rw}
\input{src/03-approach}

\input{src/04-exp}

\input{src/05-conc}
\input{src/06-ack}
{\small
\bibliographystyle{ieee_fullname}
\bibliography{egbib}
}

\end{document}

%% file: macros.tex
\definecolor{darkred}{rgb}{0.7,0.1,0.1}
\definecolor{darkgreen}{rgb}{0.1,0.7,0.1}
\definecolor{cyan}{rgb}{0.7,0.0,0.7}
\definecolor{dblue}{rgb}{0.2,0.2,0.8}
\definecolor{maroon}{rgb}{0.76,.13,.28}
\definecolor{burntorange}{rgb}{0.81,.33,0}
\definecolor{tealblue}{rgb}{0.212,0.459, 0.533}

\definecolor{mypink}{rgb}{0.93359375, 0.62109375, 0.83984375}

\definecolor{pp}{rgb}{0.43921569, 0.18823529, 0.62745098}
\definecolor{rr}{rgb}{0.5254902 , 0.00784314, 0.12941176}
\definecolor{bb}{rgb}{0.09019608, 0.23529412, 0.37647059}
\definecolor{yy}{rgb}{0.49803922, 0.3372549 , 0.0}
\definecolor{gg}{rgb}{0.02352941, 0.3372549 , 0.17647059}

\ifdefined\ShowNotes
  \newcommand{\colornote}[3]{{\color{#1}\bf{#2: #3}\normalfont}}
\else
  \newcommand{\colornote}[3]{}
\fi
\newcommand{\bb}[1]{{\textbf{\color{blue}[BB: #1]}}}
\newcommand{\jj}[1]{{\textbf{\color{cyan}[JJ: #1]}}}

\newcommand{\eat}[1]{} 
\newcommand{\name}{{\textit{RGB2Point}}}
\newcommand{\shapenetCD}{{\textit{39.26\%}}}
\newcommand{\shapenetEMD}{{\textit{26.95\%}}}
\newcommand{\pixCD}{{\textit{51.15\%}}}
\newcommand{\pixEMD}{{\textit{36.17\%}}}

\newcommand{\DNRON}{\bb{$\Downarrow$ In progress. Do not read. $\Downarrow$}}
\newcommand{\DNROFF}{\bb{$\Uparrow$ In progress. Do not read. $\Uparrow$ }}

%% file: src/00-abs.tex
\begin{abstract}
We introduce \name, an unposed single-view RGB image to a 3D point cloud generation based on Transformer. \name\ takes an input image of an object and generates a dense 3D point cloud. Contrary to prior works based on CNN layers and diffusion-denoising approaches, we use pre-trained Transformer layers that are fast and generate high-quality point clouds with consistent quality over available categories. Our generated point clouds demonstrate high quality on a real-world dataset, as evidenced by improved Chamfer distance (\pixCD) and Earth Mover's distance (\pixEMD) metrics compared to the current state-of-the-art. Additionally, our approach shows a better quality on a synthetic dataset, achieving better Chamfer distance (\shapenetCD), Earth Mover's distance (\shapenetEMD), and F-score (47.16\%). Moreover, our method produces 63.1\% more consistent high-quality results across various object categories compared to prior works. Furthermore, \name\ is computationally efficient, requiring only 2.3GB of VRAM to reconstruct a 3D point cloud from a single RGB image, and our implementation generates the results 15,133$\times$ faster than a SOTA diffusion-based model. 
\end{abstract}

%% file: src/01-intro.tex
\input{figures/overview}
\section{Introduction}
Generation of 3D point clouds from a single image is an open problem in Computer Vision, and the main challenge is handling occlusions given the limited viewpoint. The emergence of Deep Learning has alleviated this concern by leveraging 2D image features extracted from well-trained models~\cite{resnet, vgg, efficientnet} on extensive image datasets~\cite{imagenet, mscoco, cifar}. Recent works used the pre-trained models~\cite{resnet, vgg, efficientnet} as image feature extractors to reconstruct 3D objects in works~\cite{3dr2n2, pix2vox, xie2020pix2vox++, voit}. However, the introduction of the attention~\cite{transformer} mechanism and its usage in the Vision Transformer (ViT) model~\cite{vit} has shown remarkable performance improvements in image classification tasks, particularly on the ImageNet~\cite{imagenet}. ViT's unique architecture effectively captures global information through its self-attention mechanism, thus outperforming Convolutional Neural Networks (CNNs). 

Many 3D object representations exist, and unstructured point clouds are primarily provided during the data acquisition tasks as they are provided by sensors, such as LiDAR~\cite{raj2020survey}, or by photogrammetric algorithms, such as the Structure from Motion~\cite{brostow2008segmentation, schonberger2016structure}. However, they have also been used as intermediate representations in many tasks~\cite{3dr2n2, qi2017pointnet, yu2022point}. 

The denoising diffusion probabilistic model showed excellent results on 2D image synthesis~\cite{ho2020denoising, rombach2022high} and 3D objects~\cite{liu2023zero,qian2023magic123}. Many diffusion-based models require extensive hardware resources due to their large size and the numerous iterations needed to transform a Gaussian distribution into a complex one during training. This demand for resources renders some models inaccessible because of their scale and the terabytes of data required for training on internet-scale image datasets. It is widely acknowledged that gathering large volumes of data is crucial for developing stable networks. However, not all data is easy to collect, and point cloud data, particularly, have significant challenges. Collecting point cloud data requires a sensor to scan entire objects and ensure complete area coverage to avoid data gaps. This process becomes even more complicated when the target object is large or inaccessible, thus significantly increasing the complexity of data collection. Instead of relying on hardware sensors, point clouds can be obtained using photogrammetry, such as COLMAP~\cite{schoenberger2016mvs,schoenberger2016sfm}, which is grounded in Structure-from-Motion and Multi-View Stereo. Although this software-based approach offers a viable alternative, it is important to note that the quality of the generated point clouds may not be on par with those obtained from dedicated hardware sensors~\cite{nichol2022point}.

These challenges motivated us to develop a model that can reconstruct a point cloud from a single image while being executable on widely available GPUs. Our \name\ is an accessible solution that addresses the limitations inherent in existing diffusion-based models. This ensures that the broader research community can effectively utilize our methodology, facilitating wider adoption and application.

We introduce a novel approach to reconstructing images into point clouds using the ViT. Our \name\ is a Deep Learning network for 3D point cloud reconstruction from a single image. As an image feature extractor, we employ the pre-trained Vision Transformer~\cite{vit} on ImageNet~\cite{imagenet}. Our network incorporates a Multi-head Attention (MHA) layer along with a Multi-Layer Perceptron (MLP) to generate a 3D point cloud, as illustrated in~\cref{fig:overview}. Unlike heavily nested layers of models, our simple but powerful model provides high-quality but stable reconstructions over the categories. This simple architecture is cheap to train on a single desktop-level GPU and requires only 2GB of VRAM to reconstruct a 3D point cloud from a single RGB image. Besides the low memory requirements, it boosts speed as it takes 0.15 seconds per single RGB image to reconstruct a 3D point cloud.

We compare \name\ to previous approaches~\cite{baseline, ulsp, differ, xie2020pix2vox++, melas2023pc2} and demonstrate its performance through evaluation on unseen objects from two datasets: ShapeNet~\cite{shapenet}, a synthetic dataset, and Pix3D~\cite{pix3d}, a manually 3D scanned dataset from the real-world.

We analyze the impact of pre-trained weights on the effectiveness of the ViT for 3D point cloud reconstruction tasks. Moreover, we swap the image feature extractor part to a pre-trained ResNet50~\cite{resnet} to evaluate the effect of ViT.
Our method shows reconstruction improvements, including quantitative and qualitative results over existing works, as \name\ outperforms in Chamfer distance (\shapenetCD\ and \pixCD) and Earth Mover's distance (\shapenetEMD\ and \pixEMD) compared to the current state-of-the-art. Also, our model shows 47.1\% higher reconstruction quality in F-score compared to the diffusion-based method~\cite{melas2023pc2}. Contrary to previous work, our reconstructed point clouds show a 63.1\% more consistently high quality for all object classes. 
{\bf\noindent{We claim the following contributions:}}
\begin{itemize}
\item We propose a new model architecture that is VRAM efficient but also generates a high-quality point cloud from a single RGB image.
\item We show Transformer model can generate higher quality 3D objects than a probabilistic denoising diffusion model. 
\end{itemize}

%% file: figures/overview.tex
\begin{figure*}[h]
\vspace{-0.5cm}
\centering
\includegraphics[width=1.9\columnwidth]{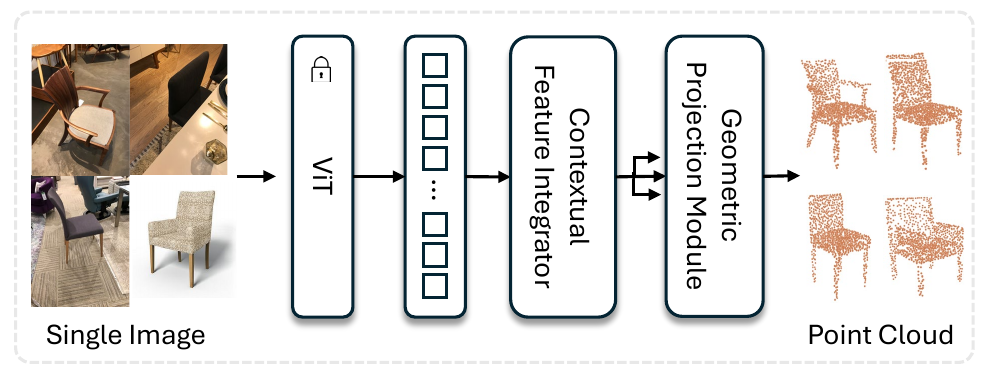}
\caption{\textbf{Model Architecture.} \name\ takes a single view RGB image and extracts image features from the pre-trained ViT~\cite{vit}. The Contextual Feature Integrator then refines these extracted features, which applies a multi-head attention mechanism~\cite{vaswani2017attention} to highlight specific regions of interest within the features. The weighted features are forwarded to the Geometric Projection Module, which maps them into a 3D space, resulting in a point cloud representation. We carefully designed the model, \name\, which requires only 2.3GB of VRAM to generate a 3D point cloud from a single RGB image.}
\label{fig:overview}
\end{figure*}

%% file: src/02-rw.tex
\section{Related Work}
\textbf{CNN based extractors} such as pre-trained models~\cite{resnet, vgg} are widely used to reconstruct 3D objects from 2D images such as an occupancy voxel~\cite{pix2vox, voit}, mesh~\cite{wang2018pixel2mesh, tang2021skeletonnet}, or 3D point cloud~\cite{3dr2n2, baseline}.
Using extracted CNN-based image features, a network learns camera poses with point cloud data to use a differentiable renderer~\cite{ulsp}. Our work does not need camera poses for point cloud generation to reduce extra efforts to maintain camera parameters for every image. Several algorithms reconstruct 3D point clouds using layers of CNN as encoder and decoder~\cite{fan2017point} and Recurrent Neural Network~\cite{3dr2n2}. An extra loss function is introduced by calculating the similarity of the reconstructed point cloud of random viewpoints~\cite{baseline}. A differentiable render is used for the 3D point cloud reconstruction~\cite{differ} to capture view consistent 3D objects.

\textbf{Transformer} uses the attention~\cite{transformer} mechanism and it has shown its outstanding performance in Natural Language Processing (NLP) such as question answering~\cite{bert, yasunaga2022linkbert, yamada-etal-2020-luke}, text generation~\cite{brown2020language, touvron2023llama} and sentiment analysis~\cite{liu2019roberta, yang2019xlnet}. Transformer~\cite{transformer} has also shown superior performances in the Computer Vision area as it outperforms image classification tasks~\cite{vit, hou2022vision}, object detection~\cite{carion2020end, fang2021you}, semantic segmentation~\cite{liu2021swin, xie2021segformer}. Moreover, the performance of Transformer~\cite{transformer} continues in 3D space as well in different domains such as 3D point cloud completion task~\cite{zhao2021point, yu2022point}, 3D reconstruction~\cite{wang2021multi} and 3D tree generation~\cite{lee2023latent}. We utilize the Vision Transformer as an image feature extractor of our model to reconstruct 3D point cloud data. Our approach exhibits improvements across various metrics, including \shapenetCD\ and \pixCD\ for Chamfer Distance, as well as \shapenetEMD\ and \pixEMD\ for Earth Mover's distance, evaluated on synthetic object datasets~\cite{shapenet} and real object datasets~\cite{pix3d}. In contrast to the existing methods~\cite{baseline, differ, ulsp} that rely on CNN-based feature extractors, our model demonstrates a better performance in 3D generation metrics as we shown in~\cref{tab:shapenet_table} from the synthetic dataset~\cite{shapenet} and~\cref{tab:pix3d_table} from the real-world dataset~\cite{pix3d}.

\textbf{Diffusion} models are new proposing approaches for a generative deep learning model by iterative denoising process. Using denoising diffusion probabilistic models (DDPM)~\cite{ho2020denoising} or from a latent space~\cite{rombach2022high}. By utilizing these two fundamental models, they contribute numerous applications such as a novel view synthesis~\cite{chan2023generative, zheng2024free3d} and a 3D object generation~\cite{liu2023zero, qian2023magic123, lee2024tree, hong2023lrm, tochilkin2024triposr}. 

In this emergence of diffusion-based models, researchers leverage its approach to point cloud such as an unconditional diffusion model to generate point clouds~\cite{luo2021diffusion, nichol2022point, vahdat2022lion}, point cloud completion using a diffusion model~\cite{lyu2021conditional} and a single image to point cloud reconstruction~\cite{melas2023pc2}. The diffusion methods leverage probabilistic approaches that show a high quality of the unseen field. 
When the number of reconstructing categories is diverse, one of the critical elements of the model is stability over the categories. From Tab.~\ref{tab:pix3d_table}, we infer the stability of each model by calculating a standard deviation of F-scores over the reconstructed categories, and the lower value gives a more stable generation quality. Also, we show \textit{15,133} times faster generation along with better qualitative results by comparing a diffusion-based work~\cite{nichol2022point} in Fig.~\ref{fig:pix3d_output}. 

Furthermore, our paper shows a Transformer-based model can also generate high quality of the unknown area like diffusion, and we show \textit{47.1\%} higher F-score than the diffusion-based method~\cite{melas2023pc2} as we show in Tab.~\ref{tab:f1score_table} and visual comparisons in~\cref{fig:pix3d_output}.

%% file: src/03-approach.tex
\section{Approach}
\textbf{Overview:} \name\ consists of three parts (\cref{fig:overview}): 1)~2D image feature extraction using a pre-trained Transformer, 2)~Contextual Feature Integrator, and 3)~Geometric Projection Module. The demonstrated strength of the transformers in vision tasks~\cite{vit, hou2022vision, swin} motivates us to integrate it for the generation of point clouds. 

The primary contribution of our work is more efficient (requires 2.3GB VRAM), faster (15,133 times faster than a diffusion-based model~\cite{nichol2022point}), higher-quality reconstructions than prior works~\cite{baseline,ulsp, melas2023pc2, nichol2022point}. A prior work~\cite{nichol2022point} uses several millions of 3D objects. \name\ requires only less than 10\% of the training dataset, and it gives a better reconstruction quality on complex real-world data as shown in~\cref{fig:pix3d_output}.

\textbf{Architecture:}
\name\ takes a $224 \times 224$ RGB image as its input and generates its corresponding 3D point cloud with $N$ points. From the evaluation, Sect.~\ref{sec:eval}, we show that our model is flexible on the number of output point clouds, such as 256, 1024, or 8192 points. We do not need any additional set of layers but simply change the size of the output layer during the training. \name\ consists of three parts. The first part is composed of a pre-trained vision Transformer~\cite{vit} that extracts image features. 

\noindent \textbf{Contextual Feature Integrator (CFI)} is a designed module to enhance the representation of specific regions within an image, which is important for generating accurate outputs. It consists of a feed-forward layer with size $A$ and a multi-head attention mechanism with $H$ heads.
The feed-forward layer acts as a transformation network, mapping the input features into a higher-dimensional space that captures relationships among the features. The dimensionality, $A$, allows the network to model complex dependencies that might not be apparent in the original feature space. By applying non-linear activation functions, this layer introduces non-linearities into the model to learn complex patterns.
The multi-head attention mechanism in the model is composed of $H$ heads, each responsible for attending to different aspects of the input feature sequences. This method enables the model to capture a wide range of contextual information within the input sequence, allowing it to attend to multiple relevant features concurrently. Each attention head computes a weighted sum of the input features, where the weights are derived from the similarity between the feature vectors, which serve as both queries, keys, and values in a self-attention setup.
The outputs from all attention heads are concatenated and passed through a linear layer, which projects them into a dimensional space appropriate for point cloud generation. This step ensures that the model effectively integrates and highlights important features, resulting in a more precise representation in the generated point cloud.

\noindent \textbf{Geometric Projection Module (GPM)} plays a key component in bridging the transition from high-dimensional feature representations to accurate 3D point cloud generation. This module maintains the spatial coherence and structural information of the input features to perform the effective translation of abstract feature spaces into meaningful geometric representations.
Following the multi-head attention mechanism, this module employs a sequence of linear layers with Leaky ReLU activations~\mbox{\cite{leaky_relu}} with a slope value of 0.2 to introduce necessary non-linearities and capture complex feature interactions. The module starts with the initial linear transformation, which projects the attention-weighted features into a higher-dimensional space. This step improves the model's capacity to capture complex spatial relationships and dependencies within the input data. Subsequent linear layers then refine these features, progressively distilling the essential information needed for accurate 3D reconstruction. The final layer of the module performs a critical projection, mapping the refined features into a three-dimensional coordinate space. This ensures that each output point corresponds to a precise $x$, $y$, and $z$ coordinate in the generated point cloud.

We set $A=1,024$, $H=4$, and $D=2,048$. We choose these values by conducting extensive experiments, as shown in~\cref{tab:ablation_shapenet_table}.

%% file: src/04-exp.tex
\section{Implementation, Datasets, and Experiments}
\label{sec:experiment}
\subsection{Implementation}
We train and test our model with Python3.9, Pytorch 2.1.0, Nvidia CUDA 12.1, a single NVIDIA RTX 4090 GPU with 24GB VRAM, and an Intel i9-13900KF.

\subsection{Datasets}
\input{figures/qualitative_results/shapenet_combined}
Since collecting RGB images with point cloud data is hard, we carefully selected our training dataset from existing data. ShapeNet~\cite{shapenet} provides a wide selection of common object categories, and since it is a synthetic dataset, we can leverage its 3D information to train our model. Working well on a synthetic dataset does not give much information for an application aspect; we evaluate our model on a complex real-world dataset, Pix2D~\cite{pix3d}, where point cloud data are scanned from the real world. We use two 3D datasets: a synthetic object dataset, ShapeNet~\cite{shapenet}, and real-world objects from Pix3D~\cite{pix3d}. We train \name\ on all available categories from ShapeNet~\cite{shapenet}. We use the same train and test splits as the previous work~\cite{baseline} that 3D-R2N2~\cite{3dr2n2} proposed. We evaluate the robustness and generalization of our model by testing it against the real-world dataset, Pix3D~\cite{pix3d}, \textit{without training our model on this dataset}.

\subsection{Training}
\name\ takes a $224 \times 224$ single RGB image from ShapeNet~\cite{shapenet}. We use the ground truth point clouds from the same datasets in resolution 1024 points, denoted by $G$. For F-score calculation (\cref{tab:f1score_table}), we simply increase the size from 1,024 to 8,192 to make fair comparisons with other works. We train our model by setting $(H=4, D=2048, A=1024)$, from~\cref{fig:overview} with a 3D point cloud reconstruction loss $L_{cd}$, that calculates Chamfer distance between the ground truth point cloud data $G$ and the generated point cloud data $R$ where 
$G=\{x_i\ \in \mathbb{R}^3\}_{i=1}^{n}, R=\{x_j\ \in \mathbb{R}^3\}_{j=1}^{n}, N(x, P) = \arg \min_{y\in P}{\|x-y\|}$. 

The model is optimized using Adam~\cite{adam} with a learning rate set to $10^{-4}$ with its default parameters, using a batch size 32. During the training, we freeze the ViT~\cite{vit} while minimizing the Chamfer distance loss $L_{cd}$, where $\alpha=5$, $n$=\textit{the number of point cloud size} and $\theta$ represents the trainable model parameters:
\begin{scriptsize}
\begin{eqnarray}
L_{cd} &= &\frac{1}{2n}\sum_{i=1}^{n}|x_{i}-N(x,R)| + \frac{1}{2n}\sum_{j=1}^{n}|x_{i}-N(x,G)|\\\label{eqn:chamferDistance}
&& \min_{\theta} \alpha L_{cd}\left(G, R, M(\theta)\right).
\label{eqn:training}
\end{eqnarray}
\end{scriptsize}

\subsection{Evaluation}\label{sec:eval}
We validated \name\ using two datasets: a synthetic dataset ShapeNet~\cite{shapenet} and real-world images from Pix3D~\cite{pix3d}. We visualize the generated results of a single RGB image from ShapeNet~\cite{shapenet} in~\cref{fig:shapenet_combined} for a qualitative evaluation. However, the qualitative results with a side-by-side comparison with some other works were not possible because of the outdated software (Python 2 version) and hardware support (CUDA 8 version) that our workstation cannot run. We contacted authors to obtain pre-trained models but could not get them from them. However, the authors provide their datasets so we can evaluate them quantitatively. Also, we compare qualitative results between a recent single image-based 3D reconstruction works~\cite{hong2023lrm, tochilkin2024triposr, nichol2022point, vahdat2022lion} in complex real-world data in~\cref{fig:pix3d_output}.
In the test stage, \name\ demonstrates efficiency, utilizing merely 2.34 GB of VRAM when operating with a batch size of one. Furthermore, a diffusion-based model~\cite{nichol2022point} takes 37 minutes and 50 seconds per image on average, but \name\ only takes 0.15 seconds per image on average. In other words, our model is 15,133 faster but generates a higher-quality point cloud in a complex real-world single image, as we show in~\cref{fig:pix3d_output}.
\input{tables/f1score}

We employ three quantitative assessment metrics to measure the similarity between the generated result and its target: Chamfer distance$\downarrow$, Earth Mover's distance$\downarrow$, and F-score$\uparrow$. The first two metrics indicate that a lower value signifies better generation quality, while the last metric shows that a higher value represents better generation quality.

First, we compare our generations to a Diffusion-based approach~\cite{melas2023pc2} and prior works that generate a 3D occupancy grid, voxel, from a single-view image~\cite{3dr2n2, yagubbayli2021legoformer, xie2020pix2vox++}. All the works are evaluated using 8192 points, and we replace the output layer size with 8,192 instead of 1,024. Based on the F-score that we show in~\cref{tab:f1score_table}, our model shows \textit{47.2\%} better than the current state-of-the-art model~\cite{melas2023pc2} that based on a probabilistic diffusion model which is recently showing high-quality 3D object generation. The SOTA method generates high-quality point clouds on specific categories, such as airplanes or rifles, but its overall stability over the categories is imbalanced. Unlike the biased performance, \name\ generates \textit{64.1\%} more stable generation quality than the SOTA method~\cite{melas2023pc2}. We calculate the performance stability using a standard deviation (Stdev.) and report it under the last row in~\cref{tab:f1score_table}. All the models are trained on the same ShapeNet~\cite{shapenet}. This shows that our model is robust and capable of generating a high-quality, dense point cloud even if we expand the size of the output layer.

Moreover, \name\ achieved Chamfer distance of $4.05\times10^2$ (car category), $5.38\times10^2$ (chair), and $2.73\times10^2$ (aircraft). These scores represent improvements of 25.00\%, 41.71\%, and 51.08\% over state-of-the-art benchmarks~\cite{differ, ulsp}. 
Additionally, employing Earth Mover's distance, our model attains $3.59\times10^2$ (car), $7.80\times10^2$ (chair), and $5.01\times10^2$ (aircraft). This indicates improvement by 24.90\%, 23.38\%, and 32.57\%, compared to the current state-of-the-art works~\cite{ulsp, differ} as summarized in~\cref{tab:shapenet_table}.

\input{tables/shapenet}

\input{figures/qualitative_results/pix3d_output}
\input{tables/pix3d}
We validate the robustness of our model on the real 3D object dataset Pix3D~\cite{pix3d}, using the trained model on ShapeNet~\cite{shapenet}. Our model generates targeted objects from the noisy background as we show generated results in~\cref{fig:pix3d_output}.
We follow the same evaluation setting as recent works~\cite{baseline, differ}. \name\ improves performs 54.57\% and 42.1\% better than the recent works~\cite{baseline, differ} in Chamfer distance, and Earth Mover's distance as shown in~\cref{tab:pix3d_table}. 

\name\ surpasses the current SOTA in performance with synthetic and real-world datasets, indicating \name\ is more robust in capturing the generated object. Using a single image, 3D Gaussian Splatting~\cite{kerbl3Dgaussians} offers a reconstruction method, albeit including the background. In contrast, our approach focuses on generating targeted objects trained amidst noisy backgrounds, as illustrated in~\cref{fig:pix3d_output}.

\subsection{Ablation Study}
We conduct five additional experiments: 1) evaluating the influence of various parameter configurations, 2) assessing how the presence of the pre-trained ViT weights affects the performance of 3D generation, 3) validating the effectiveness of ViT for point cloud generation task by replacing it to ResNet-50~\cite{resnet}, 4) evaluating the effectiveness of two modules: Contextual Feature Integrator and Geometric Projection Module, and 5) qualitative analysis of outputs using different $n$, (the number of points) where $n =$ 128.

We experimented with 16 different combinations of model parameters $(H, D, A)$ as illustrated in~\cref{fig:overview}, and evaluated the trained models using Chamfer and Earth Mover's distance (\cref{tab:ablation_shapenet_table}). The optimal parameter set is $(H=4, D=2048, A=1024)$, surpassing both \shapenetCD\ and \shapenetEMD\ metrics compared to the current state-of-the-art performance~\cite{baseline} on the ShapeNet dataset. Interestingly, our analysis reveals that more attention heads do not consistently enhance generation performance.

We also analyzed the impact of pre-trained weights on the ViT by contrasting the two-generation quality metrics obtained from models with and without pre-trained weights.
Using the models without pre-trained weights, the generations are worse with an average of 35.10\%$\pm$20.59\% and 93.11\%$\pm$15.58\% in Chamfer distance and Earth Mover's distance, respectively, compared to the models with pre-trained weights. This suggests that pre-trained weights play a crucial role in shaping 3D generations, highlighting the importance of selecting the right pre-trained weights for a particular task. 
\input{tables/ablation_shapenet}
\input{tables/ablation_no_pretrained}

Vision Transformer~\cite{vit} is widely leveraged in different computer vision tasks such as image segmentation~\cite{Xu_2022_CVPR, zhang2022segvit, gu2022multi}, depth estimation~\cite{Bhat_2021_CVPR, ranftl2021vision, zhao2022monovit} and 3D object detection~\cite{Wang_2022_CVPR}. However, we do not know the effectiveness of a single image-conditioned point cloud generation task. In this task, we evaluate the role of the image extractor by swapping out ViT to a pre-trained ResNet-50~\cite{resnet}. We train a model on the same environment to compare the generation quality on ShapeNet~\cite{shapenet} on three categories: airplane, car, and chair. As we show in~\cref{tab:ablation_cnn}, the ViT-based model generates 11.4 \% and 33.9\% better quality of point cloud given a single RGB image in terms of Chamfer Distance and Earth Mover's Distance, respectively. This study shows that similar to other computer vision tasks that leverage the power of ViT, 3D point cloud generation could be one of the fields. 
\input{tables/ablation_cnn}

Moreover, we validate the effectiveness of our two modules by training without them. We evaluate two models that remain the same but remove each module from the entire proposed architecture. We report a significant performance drop compared to metrics from the original architecture in \mbox{~\cref{tab:ablation_modules}}. Even if a module is removed, \mbox{\cref{tab:pix3d_table}} shows that our approach is at least 39.30\% and 9.88\% better in Chamfer Distance and Earth Mover's distance compared to the previous works.
\input{tables/ablation_modules}

We conduct an additional experiment by varying the number of point clouds, $n$. Specifically, we reduced the number of points to test whether this decrease affects the preservation of the object's overall shape during generation. As demonstrated in~\mbox{\cref{fig:diffrent_n}}, even with a smaller $n$, the overall shapes are still generated accurately without losing details.
\input{figures/qualitative_results/different_n}

\subsection{Limitation}
\input{figures/failed_cases}
We report failure cases using images from a real-world dataset~\cite{pix3d} in~\cref{fig:failure}. The common issue (identified in\cref{fig:failure} as (1-4)) is the lack of level of detail in our generated point cloud data. For (1), \name\ generated a chair with shorter legs than the ground truth due to the limitation of viewpoint, especially a top view in this case. Some parts of the back support are missing in the cases of (2) and (3). And (4) shows an occlusion that is about 50\% of its original shape by a desk. Compared to our failure generations, mesh reconstruction methods~\cite{hong2023lrm, tochilkin2024triposr} do not generate any related objects. Point-E~\cite{nichol2022point} cannot generate a chair at all from this real-world case scenario. LION \mbox{~\cite{vahdat2022lion}} gives better quality of point cloud than Point-E \mbox{~\cite{nichol2022point}} but still it has low accuracy based on the given RGB image as we show this behavior in \mbox{~\cref{tab:pix3d_table}}.

%% file: figures/qualitative_results/shapenet_combined.tex
\begin{figure*}[hbt]
\centering
\includegraphics[width=1.8\columnwidth]{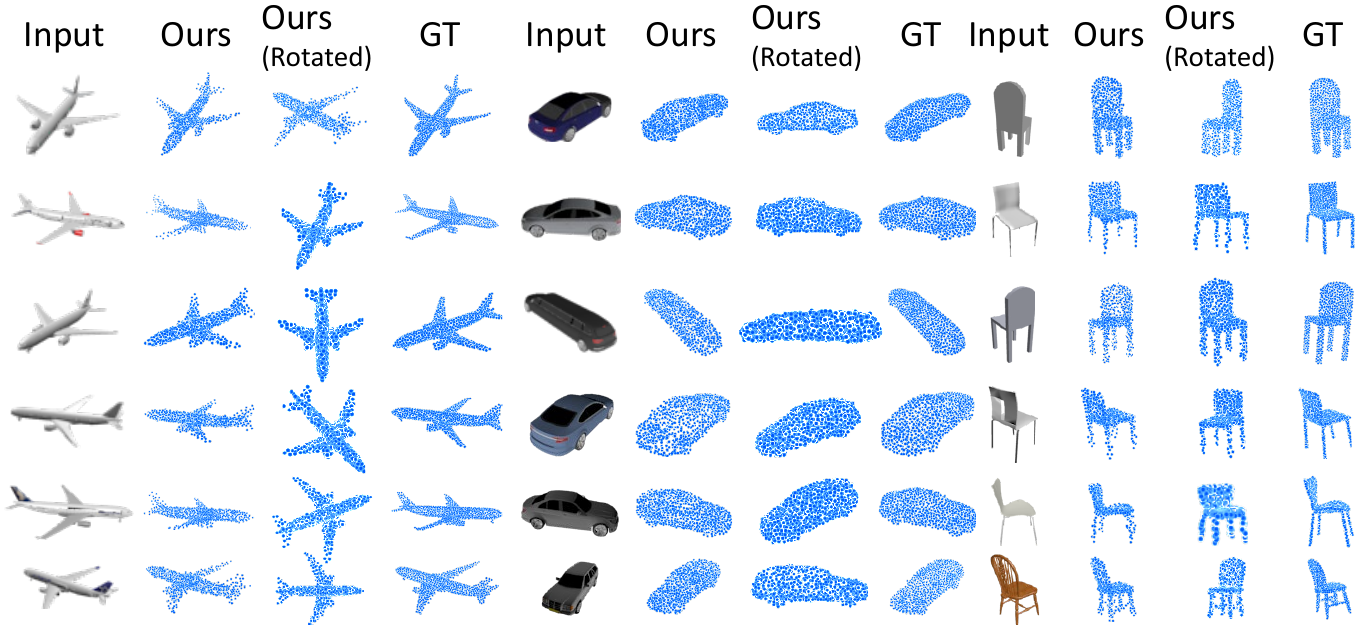}
\caption{A qualitative analysis compares 3D point clouds generated by our method, \name, from single RGB images across airplane, car, and chair categories in ShapeNet against their target point clouds.}
\label{fig:shapenet_combined}
\vspace{-0.2cm}
\end{figure*}

%% file: tables/f1score.tex
\begin{table}[hbt]
    \centering
    \small
    \begin{NiceTabular}{p{1.5cm}| p{0.5cm} p{0.5cm} p{0.5cm} p{0.5cm} p{0.5cm} p{0.5cm}}
    \specialrule{.15em}{.05em}{.05em}
    Category & ~\cite{3dr2n2} & ~\cite{yagubbayli2021legoformer} & ~\cite{xie2020pix2vox++} & ~\cite{melas2023pc2} & ~\cite{melas2023pc2}$^{2}$ &\textbf{Ours}\\ 
    \midrule
    airplane& 0.225 & 0.215 & 0.266 & 0.473 & \textbf{0.589} & \underline{0.581}\\
    bench & 0.198 & 0.241 & 0.266 & 0.305 & \underline{0.334} & \textbf{0.511}\\
    cabinet & 0.256 & 0.308 & \underline{0.317} & 0.203 & 0.211 & \textbf{0.464}\\
    car & 0.211 & 0.220 & 0.268 & 0.359 &  \underline{0.372} & \textbf{0.523}\\
    chair & 0.194 & 0.217 & 0.246 & 0.290 & \underline{0.309} & \textbf{0.544}\\
    display & 0.196 & 0.261 & \underline{0.279} & 0.232 & 0.268 & \textbf{0.487}\\
    lamp & 0.186 & 0.220 & 0.242 & 0.300 & \underline{0.326} & \textbf{0.471}\\
    loudspeaker & 0.229 & 0.286 & \underline{0.297} & 0.203 & 0.210 & \textbf{0.462}\\
    rifle & 0.356 & 0.364 & 0.410 & 0.522 & \textbf{0.585} & \underline{0.567}\\
    sofa & 0.208 & 0.260 & \underline{0.277} & 0.205 & 0.224 & \textbf{0.481}\\
    table & 0.263 & 0.305 & \underline{0.327} & 0.270 & 0.297 & \textbf{0.436}\\
    telephone & 0.407 & 0.575 & \textbf{0.582} & 0.331 & 0.389 & \underline{0.483}\\
    watercraft & 0.240 & 0.283 & 0.316 & 0.324 & \underline{0.341} & \textbf{0.552}\\
    \bottomrule
    Average$\uparrow$ & 0.244 & 0.289 & 0.315 & 0.309 & \underline{0.343} & \textbf{0.505} \\
    Stdev.$\downarrow$ & \underline{0.067} & 0.097 & 0.092 & 0.099 & 0.123 & \textbf{0.045}\\
    \specialrule{.15em}{.05em}{.05em}
    \end{NiceTabular}
    \caption{Comparison of the single image to point cloud generation on ShapeNet~\cite{3dr2n2} with prior works. We use 0.01 as a distance threshold for the F-score that higher values represent better generation quality. We \textbf{bold} the best value and \underline{underline} the current SOTA. Our model, \name, shows \textit{47.16\%} than the diffusion-based model~\cite{melas2023pc2}, which is the SOTA model. We also calculate the standard deviation (Stdev.) over the categories to evaluate a stable generation quality. \name\ shows 63.1\% stable generated point cloud quality compare to the SOTA~\cite{melas2023pc2}. \cite{melas2023pc2}$^{2}$ uses image masks to guide its generation.}  
    \label{tab:f1score_table}
\end{table}

%% file: tables/shapenet.tex
\begin{table*}[h]
    \vspace{-0.5cm}
    \small
    \centering
    \begin{NiceTabular}{p{3cm}| p{1.5cm} p{1.5cm} p{1.3cm} | p{1.5cm} p{1.5cm}p{1.3cm}}
    \specialrule{.15em}{.05em}{.05em}
      &  \multicolumn{3}{c|}{CD$(\times 10^2) \downarrow$} & \multicolumn{3}{c}{EMD$(\times 10^2)\downarrow$}\\ 
    Method &Car & Chair & Aircraft & Car & Chair & Aircraft\\ 
    \midrule
    Self-Sup.~\cite{baseline} & 10.33 & 21.84 & 15.06 & 18.32 & 23.40 & 16.12\\
    Self-Sup.~\cite{baseline}+$L_{C}$ & 6.39 & 13.58 & 8.66 & 6.42 & 16.46 & 12.53\\
    Self-Sup.~\cite{baseline}+$NN$ & 5.48 & 10.91 & 7.11 & 4.95 & 14.93 & 11.07 \\
    DIFFER~\cite{differ} & 6.35 & 9.78 & 5.67 & 6.03 & 16.21 & 9.90\\
    DIFFER~\cite{differ}+$L_{G}$ & 5.63 & \underline{9.23} & \underline{5.58} & 5.35 & 13.07 & 9.44\\
    ULSP~\cite{ulsp} & 6.64 & 10.49 & 5.70 & 6.89 & 10.93 & \underline{7.43} \\
    ULSP~\cite{ulsp}+$L_{G}$ & 6.13 & 10.0 & 7.37 & 5.83 & 10.24 & 9.99 \\
    ULSP~\cite{ulsp}+Sup. & \underline{5.4} & 9.72 & 5.91 & \underline{4.78} & \underline{10.18} & 7.66 \\
    LION~\cite{vahdat2022lion} & $-$ & 12.11 & $-$ & $-$ & 10.94 & $-$ \\
    \bottomrule
        \bf{Ours} & \bf{4.05} & \bf{5.38} & \bf{2.73} & \bf{3.59} & \bf{7.80} & \bf{5.01} \\
        Ours without CFI & 4.60 & 6.13 & 3.78 & 5.00 & 12.35 & 7.98 \\
        Ours without GPM & 5.20 & 6.59 & 4.18 & 7.45 & 11.01 & 6.97 \\
    \specialrule{.15em}{.05em}{.05em}
    \end{NiceTabular}
    \caption{The best values from different categories from ShapeNet~\cite{shapenet} in Chamfer distance (CD) and Earth Mover's distance (CMD) are noted in \textbf{bold} and the current SOTA values are \underline{underlined}. \name\ shows an average improvement of \shapenetCD\ in Chamfer distance and \shapenetEMD\ in Earth Mover's distance among the three categories.}    
    \label{tab:shapenet_table}
\end{table*}

%% file: figures/qualitative_results/pix3d_output.tex
\begin{figure}[tbh]
\centering
\includegraphics[width=0.9\columnwidth]{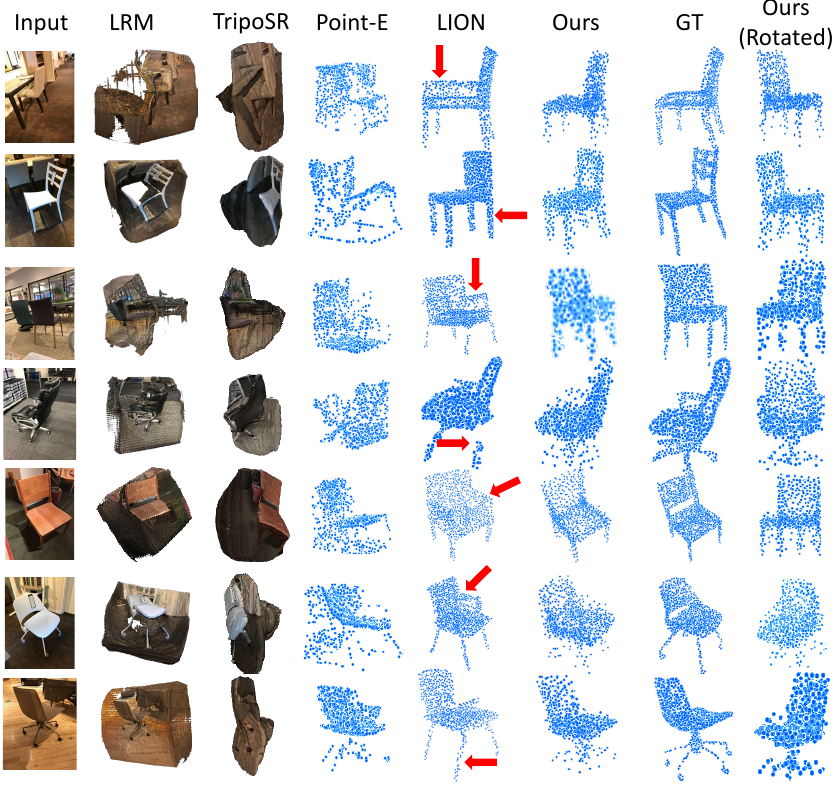}
\vspace{-0.2cm}
\caption{Generated point cloud data by \name\ using images from the real-world dataset Pix3D~\cite{pix3d}. The first column shows an input RGB image, and the next two columns show a reconstructed mesh from LRM~\cite{hong2023lrm}, TripoSR~\cite{tochilkin2024triposr}.
The third and fourth columns show reconstructed point clouds from Point-E~\cite{nichol2022point} and LION~\cite{vahdat2022lion}. The sixth left column shows generated point cloud data by \name\ and the column with \textit{GT} shows its ground truth point cloud data. The red arrows highlight differences compared to \textit{GT}. Also, we show a rotated view from our outputs in the last column.}
\label{fig:pix3d_output}
\end{figure}

%% file: tables/pix3d.tex
\begin{table}[hbt]
    \centering
    \small
    \begin{NiceTabular}{p{2.0cm}| p{2.0cm} p{2.0cm}}
    \specialrule{.15em}{.05em}{.05em}
    Method & CD$(\times 10^2)\downarrow$ & EMD$(\times 10^2)\downarrow$\\ 
    \midrule
    Self-Sup.~\cite{baseline} & 14.52 & 15.82\\
    DIFFER~\cite{differ} & \underline{14.33} & 16.09\\
    LION~\cite{vahdat2022lion} & 16.97 & \underline{14.68} \\
    \bottomrule
    \bf{Ours} & \bf{7.00} & \bf{9.37}  \\
    Ours w/o CFI & 8.53 & 11.5 \\
    Ours w/o GFM & 8.82 & 13.23 \\
    \specialrule{.15em}{.05em}{.05em}
    \end{NiceTabular}
    \caption{We test our model on the real 3D dataset Pix3D~\cite{pix3d}, and compared to the related works. The best values are shown in \textbf{bold} and the current SOTA is \underline{underlined}. \name\ shows an improvement of \pixCD\ in Chamfer distance (CD) and \pixEMD\ in Earth Mover's distance, compared to the previous state-of-the-art metrics.}  
    \label{tab:pix3d_table}
\end{table}

%% file: tables/ablation_shapenet.tex
\begin{table}[t]
    \centering
    \resizebox{0.45\textwidth}{!}{
    \begin{NiceTabular}{p{0.1cm} p{0.2cm} p{0.2cm} | p{0.5cm} p{0.8cm} p{0.8cm} | p{0.5cm} p{0.8cm} p{0.8cm} | p{0.5cm} p{0.5cm}}
    \specialrule{.15em}{.05em}{.05em}
 \multicolumn{3}{c|}{Hyper Parameters} & \multicolumn{3}{c|}{CD$(\times 10^2)\downarrow$}  & \multicolumn{3}{c|}{EMD$(\times 10^2)\downarrow$} & \multicolumn{2}{c}{Statistics$(\times 10^2)\downarrow$} \\ 
    H & D & A & Car & Aircraft & Chair & Car & Aircraft & Chair & CD Avg. & EMD Avg.\\ 
    \midrule
    2 & 1024 & 1024 & 5.46 & 2.74 & 4.05 & 8.3 & 5.43 & 3.68 & 4.08 & 5.80 \\
    2 & 1024 & 2048 & 5.48 & 2.80 & 4.10 & 8.38 & 6.03 & 3.77 & 4.13 & 6.06 \\
    2 & 1024 & 4096 & 5.41 & 2.78 & 4.12 & 8.45 & 5.69 & 3.70 & 4.10 & 5.95 \\
    2 & 2048 & 1024 & 5.44 & 2.72 & 4.06 & 8.09 & 5.16 & 3.48 & 4.07 & 5.58 \\
    2 & 2048 & 2048 & 5.40 & 2.74 & 4.12 & 8.32 & 5.74 & 3.64 & 4.09 & 5.90 \\
    2 & 2048 & 4096 & 5.48 & 2.82 & 4.09 & 8.40 & 6.03 & 3.66 & 4.13 & 6.03 \\
    4 & 2048 & 2048 & 5.47 & 2.78 & 4.12 & 8.28 & 5.42 & 3.77 & 4.12 & 5.82 \\
    \underline{4} & \underline{2048} & \underline{1024} & \underline{5.38} & \underline{2.73} & \underline{4.05} & \underline{7.80} & \underline{5.01} & \underline{3.59} & \underline{4.06} & \underline{5.47} \\
    4 & 2048 & 4096 & 5.47 & 2.80 & 4.11 & 8.51 & 5.57 & 3.70 & 4.13 & 5.93 \\
    8 & 2048 & 2048 & 5.49 & 2.75 & 4.09 & 8.50 & 5.29 & 3.73 & 4.11 & 5.84 \\
    8 & 2048 & 4096 & 5.45 & 2.84 & 4.14 & 8.28 & 5.71 & 3.80 & 4.14 & 5.93 \\
    8 & 2048 & 1024 & 5.42 & 2.77 & 4.06 & 8.09 & 5.49 & 3.55 & 4.09 & 5.71 \\
    16 & 1024 & 1024 & 5.41 & 2.74 & 4.02 & 8.29 & 6.01 & 3.53 & 4.06 & 5.94 \\
    16 & 1024 & 2048 & 5.38 & 2.78 & 4.08 & 8.30 & 5.82 & 3.64 & 4.08 & 5.92 \\
    16 & 2048 & 1024 & 5.41 & 2.76 & 4.06 & 8.04 & 5.21 & 3.55 & 4.07 & 5.60 \\
    16 & 2048 & 2048 & 5.40 & 2.77 & 4.13 & 8.10 & 5.23 & 3.67 & 4.10 & 5.67 \\
    \specialrule{.15em}{.05em}{.05em}
    \end{NiceTabular}
    }
    \caption{The ablation study using ShapeNet~\cite{shapenet} with different numbers of attention heads, \textit{H} and the dimensions of the feedforward \textit{D}, aggregator, \textit{A} from~\cref{fig:overview}. We evaluate various parameters using Chamfer Distance (CD) and Earth Mover's Distance (EMD). The best hyper-parameter set is \underline{underlined}.}    
    \label{tab:ablation_shapenet_table}

\end{table}

%% file: tables/ablation_no_pretrained.tex
\begin{table}[ht]
    \centering
    \resizebox{0.45\textwidth}{!}{
    \begin{NiceTabular}{p{0.1cm} p{0.2cm} p{0.2cm} | p{0.5cm} p{0.8cm} p{0.8cm} | p{0.5cm} p{0.8cm} p{0.8cm} | p{0.5cm} p{0.5cm}}
    \specialrule{.15em}{.05em}{.05em}
 \multicolumn{3}{c|}{Hyper Parameters} & \multicolumn{3}{c|}{CD$(\times 10^2)\downarrow$}  & \multicolumn{3}{c|}{EMD$(\times 10^2)\downarrow$} & \multicolumn{2}{c}{Difference(\%)}\\ 
    H & D & A & Car & Aircraft & Chair & Car & Aircraft & Chair & CD & EMD\\ 
    \midrule
    2 & 1024 & 1024 & 6.93 & 4.07 & 5.54 & 14.23 & 9.78 & 12.80 & 35.11 & 111.49 \\
    2 & 1024 & 2048 & 6.94 & 3.33 & 5.14 & 14.18 & 9.13 & 10.85 & 24.37 & 87.87 \\
    2 & 1024 & 4096 & 7.02 & 3.26 & 6.09 & 14.06 & 8.74 & 13.80 & 33.04 & 105.05 \\
    2 & 2048 & 1024 & 7.26 & 3.40 & 4.88 & 14.08 & 9.35 & 9.51 & 27.29 & 96.76 \\
    2 & 2048 & 2048 & 6.92 & 3.59 & 4.94 & 13.91 & 9.26 & 10.03 & 25.91 & 87.56 \\
    2 & 2048 & 4096 & 10.64 & 6.99 & 7.79 & 11.33 & 6.84 & 10.83 & 105.14 & 60.36 \\
    4 & 2048 & 2048 & 6.96 & 3.67 & 5.23 & 13.82 & 9.42 & 11.99 & 28.30 & 101.79 \\
    4 & 2048 & 1024 & 6.92 & 3.33 & 5.24 & 14.18 & 8.99 & 11.38 & 27.18 & 110.52 \\
    4 & 2048 & 4096 & 8.41 & 5.48 & 5.98 & 11.25 & 6.60 & 10.75 & 60.33 & 60.78 \\
    8 & 2048 & 2048 & 6.84 & 3.36 & 5.21 & 13.78 & 9.20 & 11.63 & 25.05 & 97.55 \\
    8 & 2048 & 4096 & 6.86 & 3.24 & 5.70 & 14.04 & 9.09 & 12.98 & 27.19 & 102.98 \\
    8 & 2048 & 1024 & 6.91 & 3.87 & 4.88 & 13.99 & 9.46 & 9.33 & 27.62 & 91.38 \\
    16 & 1024 & 1024 & 7.32 & 3.34 & 4.83 & 13.92 & 9.04 & 8.78 & 27.16 & 78.13 \\
    16 & 1024 & 2048 & 6.89 & 4.39 & 5.27 & 13.82 & 10.01 & 12.07 & 35.14 & 102.16 \\
    16 & 2048 & 1024 & 6.84 & 3.67 & 5.16 & 13.83 & 9.40 & 11.31 & 28.31 & 105.58 \\
    16 & 2048 & 2048 & 7.07 & 3.36 & 4.87 & 13.81 & 9.13 & 9.33 & 24.41 & 89.76 \\
    \specialrule{.15em}{.05em}{.05em}
    \end{NiceTabular}
    }
    \caption{We show the generation results from ShapeNet~\cite{shapenet} \textit{without} pre-trained Vision Transformer weights in Chamfer distance (CD) and Earth Mover's distance (EMD) using different parameters including attention heads, \textit{H}, dimensions of feedforward, \textit{D} and aggregator, \textit{A}. The last two columns represent performance differences depending on the existence of pre-trained weights.}
    \label{tab:ablation_shapenet_table_not_pretrained}
\end{table}

%% file: tables/ablation_cnn.tex
\begin{table}[ht]
    \centering
    \small
    \resizebox{0.45\textwidth}{!}{
    \begin{NiceTabular}{p{1.5cm} | p{0.5cm} p{0.8cm} p{0.8cm} | p{0.5cm} p{0.8cm} p{0.8cm} | p{0.8cm} p{0.8cm}}
    \specialrule{.15em}{.05em}{.05em}
 \multicolumn{1}{c|}{Model} & \multicolumn{3}{c|}{CD$(\times 10^2)\downarrow$}  & \multicolumn{3}{c}{EMD$(\times 10^2)\downarrow$}& \multicolumn{2}{c}{Average$(\times 10^2)\downarrow$}\\ 
    & Car & Aircraft & Chair & Car & Aircraft & Chair & CD & EMD \\ 
    \midrule
    ViT& 2.73 & 4.05 & 5.38 & 7.80 & 5.01 & 3.59 & 4.05 & 5.47 \\
    ResNet50& 4.55 & 3.11 & 6.06 & 5.08 & 7.57 & 12.17 & 4.57 & 8.27 \\
    \specialrule{.15em}{.05em}{.05em}
    \end{NiceTabular}
    }
    \caption{We show single image conditioned 3D point cloud generated results using two different image extractor models: ViT~\cite{vit} and ResNet50~\cite{resnet}. Both models are trained on the same environment but the only difference is its image feature extractor. We validate them using Chamfer Distance and Earth Mover's Distance on their generated point cloud data. }
    \label{tab:ablation_cnn}
\end{table}

%% file: tables/ablation_modules.tex
\begin{table}[ht]
    \centering
    \small
    \resizebox{0.45\textwidth}{!}{
    \begin{NiceTabular}{p{0.8cm} | p{1.2cm} p{1.2cm} | p{1.2cm} p{1.2cm} }
    \specialrule{.15em}{.05em}{.05em}
    
 \multicolumn{1}{c|}{Module} & \multicolumn{2}{c|}{ShapeNet}  & \multicolumn{2}{c}{Pix3D}\\ 
    & CD & EMD& CD & EMD  \\ 
    \midrule
    CFI& -19.27\%  & -54.43\%  & -21.86\%  & -22.73\% \\
    GPM& -31.30\%  & -55.06\%  & -26.00\%  & -41.20\% \\
    \specialrule{.15em}{.05em}{.05em}
    \end{NiceTabular}
    }
    \caption{We validated the effectiveness of our two modules by removing them from the model architecture, and we show the average of all scenes from this paper. The performances were measured using Chamfer Distance and Earth Mover's Distances. When we removed the modules CFI and GPM, the performance \textbf{dropped} compared to metrics from the original pipeline.}
    \label{tab:ablation_modules}
\end{table}

%% file: figures/qualitative_results/different_n.tex
\begin{figure}[tbh]
\centering
\includegraphics[width=0.95\columnwidth]{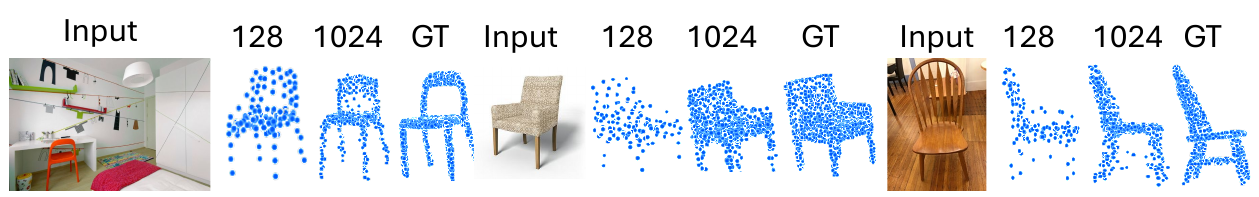}
\caption{We compare the output of different numbers of point clouds. Our original pipeline generates 1,024 point clouds but we show 128 point clouds. The overall shape is preserved instead of missing a random region of point clouds.}
\label{fig:diffrent_n}
\end{figure}

%% file: figures/failed_cases.tex
\begin{figure}[hbt]
\centering
\includegraphics[width=0.99\columnwidth]{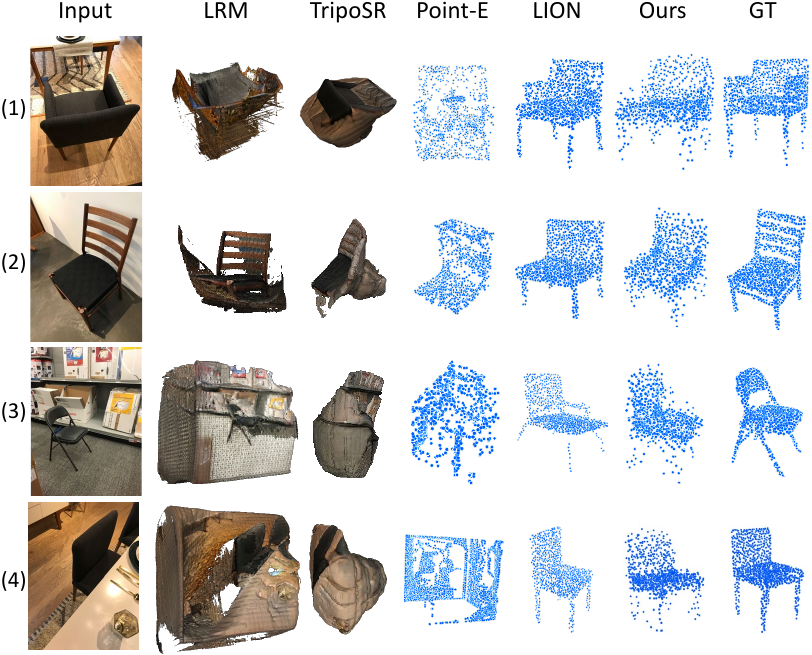}
\caption{Three failure cases from a complex real-world dataset,~\cite{pix3d},  with their input images, single image-based 3D reconstructions as a mesh~\cite{hong2023lrm, tochilkin2024triposr} and 3D point cloud~\cite{nichol2022point, vahdat2022lion}, ours, and the ground truth 3D point cloud data.}
\label{fig:failure}
\end{figure}

%% file: src/05-conc.tex
\section{Conclusions and Future Works}
We introduce a fast, high-quality 3D point cloud generative model from a single image. Leveraging the synthetic dataset from Shapenet~\cite{shapenet}, \name\ surpassed both Chamfer distance and Earth Mover's distance by \shapenetCD\ and \shapenetEMD\ respectively. In addition, our work shows an improvement on the real-world dataset~\cite{pix3d} outperforming the current SOTA by \pixCD\ and \pixEMD\ using Chamfer distance and Earth Mover's distance. In addition to the higher quality generated point cloud, our model shows a \textbf{15,133} times faster inference time than eye-catching diffusion-based models~\cite{nichol2022point, melas2023pc2}.

We explore the significance of utilizing pre-trained weights for the ViT model, showing performance on average disparities of 35.10\%$\pm$20.59\% and 93.11\%$\pm$15.58\% using Chamfer distance and Earth Mover's distance, respectively. We conclude that a pre-training weight on a 2D image affects the performance of 3D generation quality.  

Furthermore, we first, as we know of, evaluate the effectiveness of ViT compared to CNN-based image feature extractor in the 3D point cloud generation field. ViT-based generation model provides 11.4\% or 33.9\% better quality to Chamfer Distance and Earth Mover's Distance.

Also, we validate the effectiveness of our modules, Context Feature Integrator and Geometric Projection Module, in~\cref{tab:ablation_modules} that shows an average of 23.93\% and 31.97\% \textbf{performance drop} using the real datset~\cite{pix3d} in Chamfer Distance and Earth Mover's Distance.

Based on the performance and efficiency of our model, it can be used as a prior before getting actual lidar-sensor scanning, which requires multi-view scans. Our method generates a high-quality 3D point cloud from a single image in just 0.1 seconds, offering a fast and accurate alternative.

Future work could adapt \name\ for generating domain-specific objects by combining it with pre-trained weights, enabling 3D point cloud generation on desktop-level hardware. Expanding to multi-view images, with cross-attention mechanisms, could improve accuracy by leveraging complementary information from different perspectives, enhancing fidelity and robustness. Additionally, integrating a differentiable renderer for RGB texturing would increase visual quality. With VRAM efficiency, \name\ could be optimized for AR/VR deployment, achieving real-time, on-device use. Its fast generation time (0.15 seconds per image) makes it ideal for robotics tasks like path planning and object evasion.

A potential \textbf{negative societal impact} is on privacy considerations. Since our model generates a 3D point cloud from a single RGB image, a leaked photo of a new product could be used to estimate its dimensions or create a design template with minimal effort. For instance, if a new chair design is leaked before its official launch, others might attempt to produce a replicated version.

%% file: src/06-ack.tex
\section*{Acknowledgement}
This work was supported in part by the U.S. National Science Foundation under awards No. 2412928 and 2417510. Any opinions, findings, and conclusions or recommendations expressed in this material are those of the author(s) and do not necessarily reflect those of the National Science Foundation. This work is based upon efforts supported by the USDA-NIFA grants 2024-67021-42879 ``Improving Forest Management Through Automated Measurement of Tree Geometry'' and 2023-08563 ``Optimizing soybean canopy architecture for efficient light and water use under climate change''. The views and conclusions contained herein are those of the authors and should not be interpreted as representing the official policies, either expressed or implied, of the U.S. Government or NRCS. The U.S. Government is authorized to reproduce and distribute reprints for governmental purposes notwithstanding any copyright annotation therein.